%% file: neurips_2026_w.tex
\documentclass{article}

\PassOptionsToPackage{numbers, compress}{natbib}
\usepackage[sglblindworkshop, final]{neurips_2026}
\workshoptitle{RoCo-Spring: The Robust Correspondence Challenge}

\usepackage[utf8]{inputenc} % allow utf-8 input
\usepackage[T1]{fontenc}    % use 8-bit T1 fonts
\usepackage[colorlinks]{hyperref} % hyperlinks
\usepackage{url}            % simple URL typesetting
\usepackage{booktabs}       % professional-quality tables
\usepackage{amsfonts}       % blackboard math symbols
\usepackage{nicefrac}       % compact symbols for 1/2, etc.
\usepackage{microtype}      % microtypography
\usepackage[dvipsnames,table]{xcolor} % colors and table support
\usepackage{amsmath, amsthm, amssymb}
\usepackage{makecell}
\usepackage{multirow}
\usepackage{caption}
\usepackage{wrapfig}
\usepackage{graphicx}
\usepackage{subcaption}
\usepackage{algorithmic}
\usepackage{algorithm}
\usepackage{enumitem}
\usepackage[normalem]{ulem}
\hypersetup{citecolor={Brown}}
\definecolor{codebrown}{rgb}{0.37,0.0,0.0}
\definecolor{codegreen}{rgb}{0.024, 0.601, 0.012}
\input{revision-macros}

\renewcommand{\rev}[1]{#1}
\renewcommand{\revlocation}[1]{}

\newcommand{\formattedparagraph}[1]{\noindent \textbf{#1}}
\makeatletter
\newcommand{\supplementarytitlewithauthors}[2]{%
  \clearpage
  \thispagestyle{plain}
  \begingroup
  \centering
  \parindent=0pt
  \rule{\textwidth}{2.5pt}\par
  \vspace{0.1in}
  {\fontsize{17}{20}\selectfont\bfseries #1\par}
  \vspace{0.1in}
  {\large #2\par}
  \vspace{0.1in}
  \rule{\textwidth}{1.0pt}\par
  \endgroup
  \vspace{0.5in}
}
\makeatother

\title{Foundation Model Driven Optical Flow}

\author{\quad Praroop Chanda$^{1, 3}$\quad Suryansh Kumar$^{1, 2, 3, 4, *}$\\
Visual and Spatial AI Lab${^1}$, VCCM Section  \\ College of PVFA${^2}$, Department of ECEN$^3$, Department of CSCE$^4$, \\ Texas A\&M University, College Station, Texas, USA
}

\begin{document}

\maketitle
\input{sec/0-abstract-w}

\input{sec/1-intro-w}

\input{sec/3-method-w}

\input{sec/4-experiments-w}

\input{sec/5-conclusion}

% References
\bibliographystyle{abbrvnat}
\bibliography{neurips_2026.bib}

%%%%%%%%%%%%%%%%%
\newpage
\appendix
\input{sec/X-supp-w}
\input{sec/2-related-works}
\input{sec/X-supp}
\input{sec/Y-implementation}

\end{document}

%% file: revision-macros.tex
\usepackage{tikz}
\usetikzlibrary{arrows.meta,positioning,calc}
\usepackage{etoolbox}
\newif\ifshowrevisions
\showrevisionsfalse
\DeclareRobustCommand{\rev}[1]{%
  \ifshowrevisions{\begingroup\color{blue}%
    \hypersetup{citecolor=blue,linkcolor=blue,urlcolor=blue}#1\endgroup}%
  \else#1\fi}
\newenvironment{revision}{%
  \par\begingroup\ifshowrevisions\color{blue}%
  \hypersetup{citecolor=blue,linkcolor=blue,urlcolor=blue}\fi
}{\par\endgroup}
\newcommand{\revlocation}[1]{\phantomsection\label{rev:#1}}

\makeatletter
\AtBeginDocument{%
  \let\revision@lbibitem\@lbibitem
  \def\@lbibitem[#1]#2{%
    \color{black}\hypersetup{urlcolor=magenta,citecolor=Brown,linkcolor=red}%
    \ifshowrevisions
      \ifboolexpr{test {\ifstrequal{#2}{wang2025waft}} or
                   test {\ifstrequal{#2}{zhang2026d4rt}} or
                   test {\ifstrequal{#2}{fflowCode2026}} or
                   test {\ifstrequal{#2}{roco29Public2026}} or
                   test {\ifstrequal{#2}{dinov2SmallConfig}} or
                   test {\ifstrequal{#2}{spynet2017}}}%
    {\color{blue}\hypersetup{urlcolor=blue,citecolor=blue,linkcolor=blue}}{}%
    \fi
    \revision@lbibitem[#1]{#2}}%
}
\makeatother

%% file: sec/0-abstract-w.tex
\begin{abstract}
\revlocation{abstract}
\rev{We investigate how frozen pretrained visual and depth representations support dense optical flow when coupled with a supervised matching network. Building on TTSFlow, this paper studies Dense Prediction Transformer (DPT) feature decoding and one coarse-to-fine refinement stage. Our method fuses features from frozen DINOv2 and Depth Anything V2 models and estimates optical flow through a trainable transformer-based matching pipeline. After optical-flow training on FlyingChairs and FlyingThings3D, it achieves 2.37 endpoint error (EPE) on the Sintel training split, Final pass, compared with 2.52 for the FlowSeek (T) configuration reported here.}
Our RoCo-Spring 2026 entry, \textbf{RoCo-29 $\cdot$ VSAI}, obtains a robustness-balanced score of 1.1038, Spring EPE of 1.411, and RobustSpring proxy EPE of 4.739,
\rev{with a lower balanced score than seven of the twelve listed reference baselines. These results characterize the complete  pipeline, which reuses frozen foundation backbones and performs one refinement stage.}
Code is available at: \url{https://github.com/Visual-and-Spatial-AI-Lab/FFlow}
\end{abstract}

%% file: sec/1-intro-w.tex
\section{Introduction}\label{sec:intro}
\revlocation{positioning}
Dense optical flow lies at the core of visual perception \cite{sun2010secrets}. It estimates a motion vector for every pixel between consecutive frames. It enables a wide range of applications, including video understanding \cite{gao2020flow}, saliency detection \cite{li2019motion}, action recognition \cite{zhao2020improved, piergiovanni2019representation, sun2018optical}, dense 3D reconstruction \cite{ma2022multiview, kumar2019superpixel, ranftl2016dense}, and motion modeling \cite{huang2024zero}. Over the past decade, deep learning-based approaches have dramatically improved the accuracy of optical flow \cite{deng2023explicit, dong2023rethinking, garrepalli2023dift, huang2022flowformer, leroy2023win, lu2023transflow, saxena2024surprising, shi2023flowformer++, sui2022craft, sun2018pwc, sun2022skflow, teed2020raft, weinzaepfel2022croco, weinzaepfel2023croco, xu2022gmflow, zheng2022dip, zhao2022global}. \rev{Notable methods such as RAFT~\cite{teed2020raft} and its variants~\cite{wang2024sea,huang2022flowformer,Poggi_2025_ICCV} combine learned representations with repeated flow updates. This motivates studying how pretrained representations can be used within an alternative matching pipeline.}

In parallel, vision foundation models have reshaped the field of computer vision \cite{oquab2023dinov2, caron2021emerging, kirillov2023segment, radford2021learning, dosovitskiy2020image}. Models trained once on internet-scale imaging data now transfer effectively across computer vision tasks such as object detection \cite{zhuang2025argus}, segmentation \cite{zhang2024improving}, depth estimation \cite{yang2024depth, yang2024depthv2}, and geometric reasoning \cite{liu2023segment}. Rather than retraining specialized architectures for each problem, we increasingly reuse pretrained representations as general-purpose visual priors. \rev{These developments motivate reusing pretrained representations while learning the modules needed for optical-flow estimation.}

Pretrained representations have also been investigated for optical flow, including cross-view pretraining in CroCo v2~\cite{weinzaepfel2023croco}, depth-informed estimation in FlowSeek~\cite{Poggi_2025_ICCV}, and pretrained input encoders in WAFT~\cite{wang2025waft}. Our question is how to combine frozen visual and depth representations within a dense optical-flow pipeline. We evaluate that pipeline on cross-dataset optical-flow benchmarks and the RoCo-Spring corruption benchmark. Appendix~\ref{sec:relatedwork} discusses these approaches and the related video-based reconstruction setting of D4RT~\cite{zhang2026d4rt}.

\revlocation{contribution}
This paper extends our TTSFlow framework~\cite{Chanda_2026_CVPR}, retaining its combination of frozen DINOv2 and Depth Anything V2 features before matching. The changes studied here are DPT-based decoding of intermediate DINOv2 features, one higher-resolution refinement stage, and evaluation of a Spring-adapted model on RoCo-Spring. The matching transformer, correlation-to-flow formulation, feature-based propagation, coarse-to-fine refinement, and convex upsampling follow GMFlow~\cite{xu2022gmflow}. These are inherited components, not new matching operators. The downstream transformer, projection and fusion layers, propagation projections, and upsampling predictor are learned from optical-flow supervision. Experiment (Table~\ref{tab:dpt_ablation}) compares DPT configurations at a fixed refinement count. The historical TTSFlow fusion ablation is explicitly identified in the experiments (Table~\ref{tab:fusion_ablation}). Our experimental results evaluate this combined system rather than isolating what frozen foundation features can achieve alone.

%% file: sec/3-method-w.tex
\section{Method}\label{sec:methodology}
Given two consecutive RGB frames $\mathbf{I}_1,\mathbf{I}_2\in\mathbb{R}^{H\times W\times3}$, we estimate dense optical flow $\mathbf{V}\in\mathbb{R}^{H\times W\times2}$, where $H$ and $W$ denote image height and width, respectively.
\rev{We fuse two frozen foundation-model feature streams and apply supervised matching at $1/8$ resolution, followed by one residual refinement at $1/4$ resolution. Here $H,W$ refer to the dimensions supplied to the model. Figure~\ref{fig:overall_pipeline} below summarizes the overall pipeline whereas Appendix~\ref{app:implementation} records the released configuration, parameter inventory, and its gradient-status distinction.}

\revlocation{visual}
\formattedparagraph{1. Foundation Visual Feature Extraction.}\rev{
The visual branch resizes each frame to $448\times448$ by bilinear interpolation. Frozen DINOv2-S~\cite{oquab2023dinov2} uses $14\times14$ patches, giving a $32\times32$ patch-token grid with 384 channels. We take transformer-block outputs 3, 6, 9, and 12 (one-based indexing), remove their class tokens, and decode the four spatial maps with a DPT head~\cite{ranftl2021vision}. Its finest 128-channel output is resized separately to the two matching grids, i.e.,
\begin{equation}
\mathbf{F}_i^{D,s}=\mathcal{R}_{s}\!\left(\Psi_{\mathrm{DPT}}\!\left(
\{\widetilde{\mathbf{F}}_i^{D,\ell}\}_{\ell\in\{3,6,9,12\}}\right)\right),
\quad s\in\{8,4\},
\label{eq:visualpath}
\end{equation}
where $\mathcal{R}_s$ bilinearly resizes to $H/s\times W/s$. Thus, $1/8$ and $1/4$ are post-decoding matching resolutions, not native DINOv2 resolutions. We keep the pretrained encoder fixed and studied the adaptation setting.}

\revlocation{depth}
\formattedparagraph{2. Depth Prior Feature Encoding.}\rev{
The depth branch uses frozen Depth Anything V2-B~\cite{yang2024depthv2}, including its pretrained DPT decoder, with a $518\times518$ input and a $37\times37$ encoder patch grid. We use the 128-channel output of the last decoder fusion block, \texttt{depth\_head.scratch.refinenet1}, before \texttt{output\_conv1}. The scalar depth output is not used for matching. This feature is resized to $H\times W$ and projected to both scales:
\begin{equation}
\widetilde{\mathbf{F}}_i^{Z,s}=\Psi_{\mathrm{proj}}^s(\mathbf{F}_i^Z),
\qquad \widetilde{\mathbf{F}}_i^{Z,s}\in\mathbb{R}^{H/s\times W/s\times128}.
\label{eq:depthpath}
\end{equation}
Three $3\times3$, stride-2 convolutions use channel widths $128\!\rightarrow\!192\!\rightarrow\!256\!\rightarrow\!128$; the second and third outputs supply $1/4$ and $1/8$ features. A $1\times1$ projection reduces the $1/4$ output to 128 channels.
}

\revlocation{fusion}
\formattedparagraph{3. Cross-Modal Feature Fusion.}
\rev{Both branches take the same RGB frame and reflect different pretraining objectives. At each scale, we concatenate their 128-channel features to obtain $\mathbf{F}_i^C=\operatorname{Concat}(\mathbf{F}_i^D,\widetilde{\mathbf{F}}_i^Z)$, suppressing the scale index for readability.}
\begin{equation}
\hat{\mathbf{F}}_i=\Psi_{\mathrm{fusion}}(\mathbf{F}_i^C).
\label{eq:fusion}
\end{equation}
\rev{Fusion applies a $3\times3$ convolution and two residual blocks at 256 channels, then a $1\times1$ projection to 128 channels. These weights are shared across frames and scales. Fusion precedes matching so that similarity is computed on a unified representation. Appendix~\ref{app:parameters} gives exact source-derived counts rather than characterizing this network as merely lightweight.}

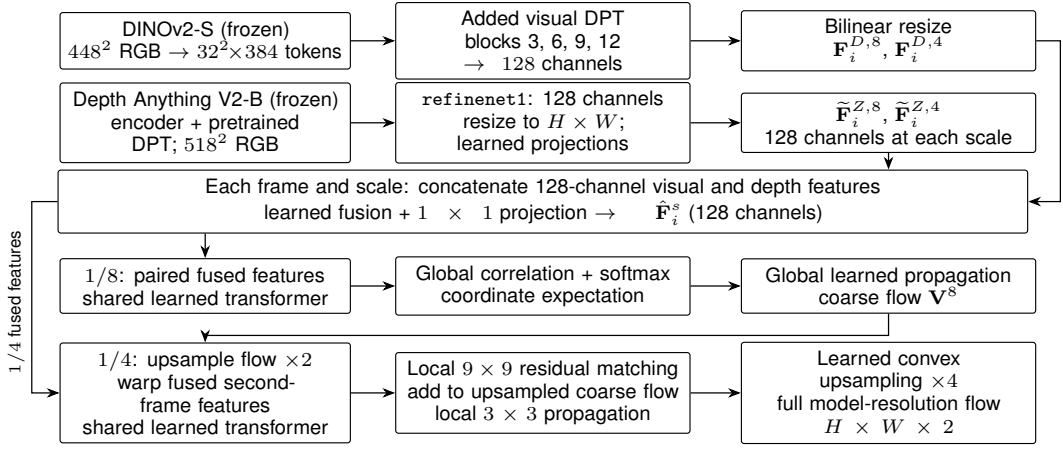
\begin{figure}[t]
\centering
\input{sec/figure1-refinement}
\caption{\rev{\textbf{Overall pipeline, including refinement.} Each feature branch is applied to both RGB frames. The TTSFlow feature-fusion design~\cite{Chanda_2026_CVPR} is extended with visual DPT decoding and two feature scales. Matching, propagation, refinement, and convex upsampling follow GMFlow~\cite{xu2022gmflow}. The transformer is shared across scales. Fusion and projection weights are shared. Frozen labels apply to the pretrained foundation models. The visual DPT gradient status in the released code is documented in Appendix~\ref{app:parameters}.}}
\label{fig:overall_pipeline}
\end{figure}

\revlocation{matching}
\formattedparagraph{4. Transformer-Based Global Matching and Propagation.}
We retain the global matching formulation of GMFlow~\cite{xu2022gmflow} while deriving features from frozen foundation backbones~\cite{oquab2023dinov2,yang2024depthv2}.
\rev{After adding positional encoding, a trainable six-block transformer processes the fused features with 128-dimensional embeddings, one attention head, and self-/cross-attention. The two scales use $2\times2$ and $8\times8$ spatial window partitions, respectively.}
We reshape its outputs as $\mathbf{F}_i\in\mathbb{R}^{N\times D}$, where $N=(H/8)(W/8)$ \rev{and $D=128$}. We compute all-pairs correlations and correspondence probabilities as
$\mathbf{C}_{\mathrm{flow}}=(\mathbf{F}_1\mathbf{F}_2^T)/\sqrt{D}$,
$\mathbf{M}_{\mathrm{flow}}=\operatorname{softmax}(\mathbf{C}_{\mathrm{flow}})$,
where both matrices lie in $\mathbb{R}^{N\times N}$. Row-wise softmax assigns each reference location a distribution over all target locations, without restricting the displacement search range.

Let $\mathbf{G}_{2\mathrm{D}}\in\mathbb{R}^{N\times2}$ denote the coordinate grid shared by both frames at feature resolution, flattened consistently with $\mathbf{F}_i$. The expected target coordinates and initial flow are $\hat{\mathbf{G}}_{2\mathrm{D}}=\mathbf{M}_{\mathrm{flow}}\mathbf{G}_{2\mathrm{D}}$,
$\hat{\mathbf{V}}_{\mathrm{flow}}=\hat{\mathbf{G}}_{2\mathrm{D}}-\mathbf{G}_{2\mathrm{D}}$.
This expectation produces continuous correspondence coordinates without a discrete argmax.
\rev{Global propagation uses learned projections $\mathbf{Q}=\psi_q(\mathbf{F}_1)$ and $\mathbf{K}=\psi_k(\mathbf{Q})$ in the released implementation, giving $\mathbf{A}=\operatorname{softmax}(\mathbf{Q}\mathbf{K}^{\top}/\sqrt D)$ and $\mathbf{V}=\mathbf{A}\hat{\mathbf{V}}_{\mathrm{flow}}$. The correlation/expectation operators are parameter-free given the transformed features; the surrounding network is learned.}

\revlocation{refinement}
\formattedparagraph{\rev{5. Refinement and Final Upsampling.}} Following GMFlow~\cite{xu2022gmflow}, we bilinearly upsample the propagated $1/8$ flow by two and multiply its displacements by two. The detached flow warps the second-frame fused $1/4$ features. The shared transformer processes the first-frame and warped second-frame features. Local matching in a $9\times9$ neighborhood predicts a residual added to the up-sampled flow. Propagation then uses a $3\times3$ neighborhood. Final learned convex upsampling uses $3\times3$ flow neighborhoods and a factor of four, also multiplying displacements by four. One refinement stage means one higher-resolution residual update after global matching, and thus not refinement-free inference.

\revlocation{loss}
\formattedparagraph{\rev{6. Training Loss.}} We supervise four full-resolution predictions, i.e., coarse matching, coarse propagation, fine residual matching, and final fine propagation with convex upsampling. The first three use bilinear upsampling with displacement scaling. The masked $\ell_1$ loss is
$\mathcal{L}_{\mathrm{flow}}=\sum_{k=1}^{4}\gamma^{4-k}
\operatorname{mean}\!\left[m\,|\mathbf{V}^{(k)}-\mathbf{V}_{gt}|\right]$,
with $\gamma=0.9$ and weights $(0.729,0.81,0.9,1)$. Our released code loss masks invalid pixels and ground-truth magnitudes at least 400 pixels, then averages over the full batch/spatial/two-component tensor. Appendix~\ref{app:supervision} states the normalization and prediction order explicitly.

%% file: sec/figure1-refinement.tex
% Kumar: Revised the figure due to Reviewer's comment.
\begingroup
\ifshowrevisions\color{blue}\hypersetup{linkcolor=blue,citecolor=blue,urlcolor=blue}\def\diagramcolor{black}\else\def\diagramcolor{black}\fi
\resizebox{\linewidth}{!}{%
\begin{tikzpicture}[x=1cm,y=1cm,
 every node/.style={font=\sffamily\scriptsize,text=\diagramcolor,align=center},
 box/.style={draw=\diagramcolor,rounded corners=1pt,inner sep=3pt,minimum height=0.63cm},
 arr/.style={-{Stealth[length=1.6mm]},draw=\diagramcolor,line width=0.45pt}]
\node[box,text width=3.4cm] (v) at (1.8,0) {DINOv2-S (frozen)\\$448^2$ RGB $\to32^2\!\times384$ tokens};
\node[box,text width=3.45cm] (dpt) at (6.0,0) {Added visual DPT\\blocks 3, 6, 9, 12 $\to128$ channels};
\node[box,text width=3.45cm] (vf) at (10.3,0) {Bilinear resize\\$\mathbf{F}^{D,8}_i$, $\mathbf{F}^{D,4}_i$};
\draw[arr] (v)--(dpt);\draw[arr] (dpt)--(vf);
\node[box,text width=3.4cm] (z) at (1.8,-1.02) {Depth Anything V2-B (frozen)\\encoder + pretrained DPT; $518^2$ RGB};
\node[box,text width=3.45cm] (zp) at (6.0,-1.02) {\texttt{refinenet1}: 128 channels\\resize to $H\!\times\!W$; learned projections};
\node[box,text width=3.45cm] (zf) at (10.3,-1.02) {$\widetilde{\mathbf{F}}^{Z,8}_i$, $\widetilde{\mathbf{F}}^{Z,4}_i$\\128 channels at each scale};
\draw[arr] (z)--(zp);\draw[arr] (zp)--(zf);
\node[box,text width=11.85cm,minimum height=0.58cm] (fusion) at (6.0,-2.00) {Each frame and scale: concatenate 128-channel visual and depth features\\learned fusion + $1\times1$ projection $\to\hat{\mathbf{F}}_i^s$ (128 channels)};
\draw[arr] (vf.east)--++(0.30,0)|-(fusion.east);
\draw[arr] (zf.south)--(zf.south|-fusion.north);
\node[box,text width=3.4cm] (t8) at (1.8,-3.05) {$1/8$: paired fused features\\shared learned transformer};
\node[box,text width=3.45cm] (m8) at (6.0,-3.05) {Global correlation + softmax\\coordinate expectation};
\node[box,text width=3.45cm] (p8) at (10.3,-3.05) {Global learned propagation\\coarse flow $\mathbf{V}^{8}$};
\draw[arr] (fusion.south-|t8.north)--(t8.north);
\draw[arr] (t8)--(m8);\draw[arr] (m8)--(p8);
\node[box,text width=3.4cm] (w4) at (1.8,-4.38) {$1/4$: upsample flow $\times2$\\warp fused second-frame features\\shared learned transformer};
\node[box,text width=3.45cm] (m4) at (6.0,-4.38) {Local $9\times9$ residual matching\\add to upsampled coarse flow\\local $3\times3$ propagation};
\node[box,text width=3.45cm] (out) at (10.3,-4.38) {Learned convex upsampling $\times4$\\full model-resolution flow\\$H\times W\times2$};
\draw[arr] (p8.south)--++(0,-0.27)-|(w4.north);
\draw[arr] (w4)--(m4);\draw[arr] (m4)--(out);
\draw[arr] (fusion.west)--++(-0.32,0) |- (w4.west);
\node[font=\sffamily\tiny,rotate=90,fill=white,inner sep=1pt] at (-0.55,-3.25) {$1/4$ fused features};
\node[font=\sffamily\tiny,align=left,anchor=west] at (0.0,-5.29) {Visual DPT: no flow gradient in the released forward path (Appendix~\ref{app:parameters}). Upsampling also scales flow magnitudes.};
\end{tikzpicture}}
\endgroup

%% file: sec/4-experiments-w.tex
\section{Experiments}
\revlocation{training}
\formattedparagraph{Implementation details.} We implement the method in PyTorch~\citep{paszke2019pytorch} and train with AdamW~\cite{loshchilov2017decoupled} on two NVIDIA RTX 6000 GPUs. Initial training uses FlyingChairs for 200k steps, learning rate $4\!\times\!10^{-4}$, and $384\!\times\!512$ crops. FlyingThings3D training uses learning rate $2\!\times\!10^{-4}$ and $384\!\times\!768$ \rev{crops} for 360k iterations~\citep{dosovitskiy2015flownet,mayer2016large}. \rev{The released launcher specifies global batch sizes 8 for Chairs/Things and 4 for TSKH, KITTI, and Spring. Its paired random-crop procedures and stage settings are provided in Appendix~\ref{app:training}.}

Cross-dataset evaluation uses Chairs and Things-trained models on the Sintel training splits without Sintel adaptation. Target-adapted evaluations follow training on the Things \cite{mayer2016large, dosovitskiy2015flownet}, Sintel \citep{butler2012naturalistic}, KITTI \citep{geiger2013vision}, HD1K mixture (TSKH) \citep{kondermann2016hci} with a learning rate of $2\!\times\!10^{-4}$ and a crop size of $320\!\times\!896$. We train our  model for 320k iterations. For KITTI, we additionally fine-tune the model for 200k iterations with a batch size 4 \rev{(see Appendix~\ref{app:training} for the crop-width discrepancy in the release)}. These training-split results are not held-out test performance. Finally, for the RoCo-Spring entry we fine tune our model on the Spring \cite{mehl2023spring} dataset for 400k iterations, using a batch size 4,  crop size of $544 \times 960$ and a learning rate of  $1 \times 10^{-4}$. \rev{The predicted optical-flow fields files in .hdf5 format are submitted to the benchmark server for evaluation.} 

\begin{table*}
\centering
\scriptsize
%\setlength{\tabcolsep}{4pt}
%\vspace{\baselineskip}
\resizebox{1.0\linewidth}{!}{
\begin{tabular}{l|c|c|cccc|cccc}
\hline
\multirow{2}{*}{Method} & \multirow{2}{*}{\#refine} 
& \multicolumn{1}{c|}{Things (val)} 
& \multicolumn{4}{c|}{Sintel (train, clean)} 
& \multicolumn{4}{c}{Sintel (train, final)} \\
\cline{3-11}
& & EPE
& EPE & s$_{0-10}$ & s$_{10-40}$ & s$_{40+}$ 
& EPE & s$_{0-10}$ & s$_{10-40}$ & s$_{40+}$ \\
\hline

RAFT \cite{teed2020raft} & 32 & 4.25 
& 1.43 & 0.33 & 1.54 & 9.04 
& 2.71 & 0.51 & 2.98 & 17.62 \\
\hline

GMFlow \cite{xu2022gmflow} & 0 & 3.48 
& 1.50 & 0.46 & 1.77 & 8.26 
& 2.96 & 0.72 & 3.45 & 17.70 \\
\hline

GMFlow \cite{xu2022gmflow} & 1 & 2.80 
& 1.08 & 0.30 & 1.25 & 6.26 
& 2.48 & 0.51 & 2.81 & 15.76 \\
\hline

SEA-RAFT (S) \cite{wang2024sea} & 4 & - 
& 1.27 & - & - & - 
& 4.32 & - & - & - \\
\hline

FlowSeek (T) \cite{Poggi_2025_ICCV} & 4 & 3.94
& 1.12 & 0.23 & 1.27 & 7.12 
& 2.52 & 0.45 & 2.78 & 16.60 \\
\hline

{TTSFlow \cite{Chanda_2026_CVPR}} & {0}
& 3.02 
& 1.46 & 0.39 & 2.05 & 7.74 
& 2.81 & 0.64 & 3.36 & 16.99 \\
\hline

\textbf{Ours } & \textbf{1}
& \textbf{3.28} 
& \textbf{1.26} & 0.34 & 1.52 & 7.25 
& \textbf{2.37} & 0.55 & 2.92 & \textbf{14.05} \\
\hline
\end{tabular}
}
\caption{\small \textbf{Generalization} performance after training on Chairs and Things train sets. \rev{The 2.37 EPE is on the Sintel training split, Final pass, and is the lowest among these listed configurations. The TTSFlow row is reproduced from~\cite{Chanda_2026_CVPR}; refinement counts are not runtime measurements.}}\label{tab:zeroshot_full}
\end{table*}

\subsection{Model Generalization} 
Table~\ref{tab:zeroshot_full} summarizes cross-dataset generalization results after training exclusively on the FlyingChairs and FlyingThings3D datasets, with no dataset-specific fine-tuning applied at test time.

\revlocation{generalization}
\rev{On FlyingThings3D validation, our EPE of 3.28 is below RAFT (4.25) and GMFlow without refinement (3.48), but above GMFlow with one refinement (2.80). On the Sintel training split, our Final-pass EPE is 2.37, compared with 2.48 for GMFlow with one refinement and 2.52 for the listed FlowSeek (T) configuration. On Clean, our EPE is 1.26, compared with 1.08 and 1.12, respectively. These results identify a Final-pass advantage within the listed configurations, while the ordering differs on Clean and Things.}

%after fine-tuning on KITTI, following training on Chairs, Things, and mixed dataset (TSKH).

%\vspace{-0.2cm}

\subsection{Result on Sintel and KITTI}

\formattedparagraph{I. Results on Sintel.}
Table~\ref{tab:sintel_trainedg} reports Sintel training-set results after training on Chairs, Things, and TSKH. TTSFlow \cite{Chanda_2026_CVPR} achieves 0.847, 1.140 EPE on Clean and Final sequence, respectively. On Final, TTSFlow \cite{Chanda_2026_CVPR} outperforms RAFT, GMFlow without refinement, and FlowSeek, while our model obtains 0.816, 1.13 EPE superseding TTSFlow \cite{Chanda_2026_CVPR} results. SEA-RAFT achieves the lowest errors on both passes, although training configurations differ: it uses eight L40 GPUs, batch size 32, and $432 \times 960$ crops, compared with two RTX 6000 GPUs, batch size 4, and $320 \times 896$ crops for ours. Both SEA-RAFT and FlowSeek also use additional TartanAir pretraining, whereas our approach reuses frozen foundation backbones without this flow-pretraining stage. Thus, these comparisons establish relative accuracy under the reported protocols, rather than isolate the effects of architecture or training scale.

\formattedparagraph{II. Results on KITTI.}
On the KITTI training set (Table~\ref{tab:kitti_trainedg}), TTSFlow \cite{Chanda_2026_CVPR} achieves 1.99 EPE and 7.40\% F1-all, slightly improving on the corresponding GMFlow configuration (2.06 EPE and 7.57\%). In comparison, our model achieves 1.248 EPE and 4.01\% F1-all, also outperforming GMFlow with one refinement (1.36 EPE and 5.17\%). \rev{RAFT and SEA-RAFT have lower errors on both metrics. FlowSeek has slightly higher EPE (1.26 versus 1.248), but lower F1-all (3.90\% versus 4.01\%).} Thus, our approach improves on GMFlow at both reported refinement counts, but the accuracy gap to the strongest KITTI baselines remains.

\begin{table}[t]
\centering
\scriptsize
\setlength{\tabcolsep}{4pt}
\begin{tabular}{@{}lcc cccc@{}}
\toprule
Method & \rev{Extra flow data} & \#refine
& \multicolumn{2}{c}{Sintel (train)}
& \multicolumn{2}{c}{KITTI (train)} \\
\cmidrule(lr){4-5}
\cmidrule(lr){6-7}
& & &
Clean EPE & Final EPE & EPE & F1-all (\%) \\
\midrule

RAFT \cite{teed2020raft}
& -- & 32
& 0.768 & 1.217
& 0.63 & 1.47 \\

GMFlow \cite{xu2022gmflow}
& -- & 0
& 0.947 & 1.276
& 2.06 & 7.57 \\

GMFlow \cite{xu2022gmflow}
& -- & 1
& 0.762 & 1.110
& 1.36 & 5.17 \\

SEA-RAFT (S) \cite{wang2024sea}
& TartanAir & 4
& 0.546 & 0.782
& 0.93 & 2.65 \\

FlowSeek (T) \cite{Poggi_2025_ICCV}
& TartanAir & 4
& 0.710 & 1.280
& 1.26 & 3.90 \\

{TTSFlow \cite{Chanda_2026_CVPR}}
& {--} & {0}
& {0.847} & {1.140}
& {1.99} & {7.40} \\

\midrule

\textbf{Ours}
& \textbf{--} & \textbf{1}
& \textbf{0.816} & \textbf{1.13}
& \textbf{1.248} & \textbf{4.01} \\

\bottomrule
\end{tabular}
\caption{\small Performance on the Sintel and KITTI training sets. Sintel results are obtained after training on Chairs, Things, and the mixed TSKH datasets. \rev{Training configurations differ; the comparison does not isolate architecture or training-scale effects. Extra flow data denotes additional optical-flow pretraining, not foundation-model pretraining. The TTSFlow row is reproduced from~\cite{Chanda_2026_CVPR}.} \#refine denotes the number of refinement steps. Lower values indicate better performance for all metrics.}
\label{tab:sintel_trainedg}
\label{tab:kitti_trainedg}
\end{table}

%Ablation study evaluating the 

%\vspace{-0.3cm}
%\vspace{-0.2cm}

%
\subsection{RoCo-Spring Challenge Benchmark}
\begin{figure*}[t]
\centering
\includegraphics[width=0.9\linewidth]{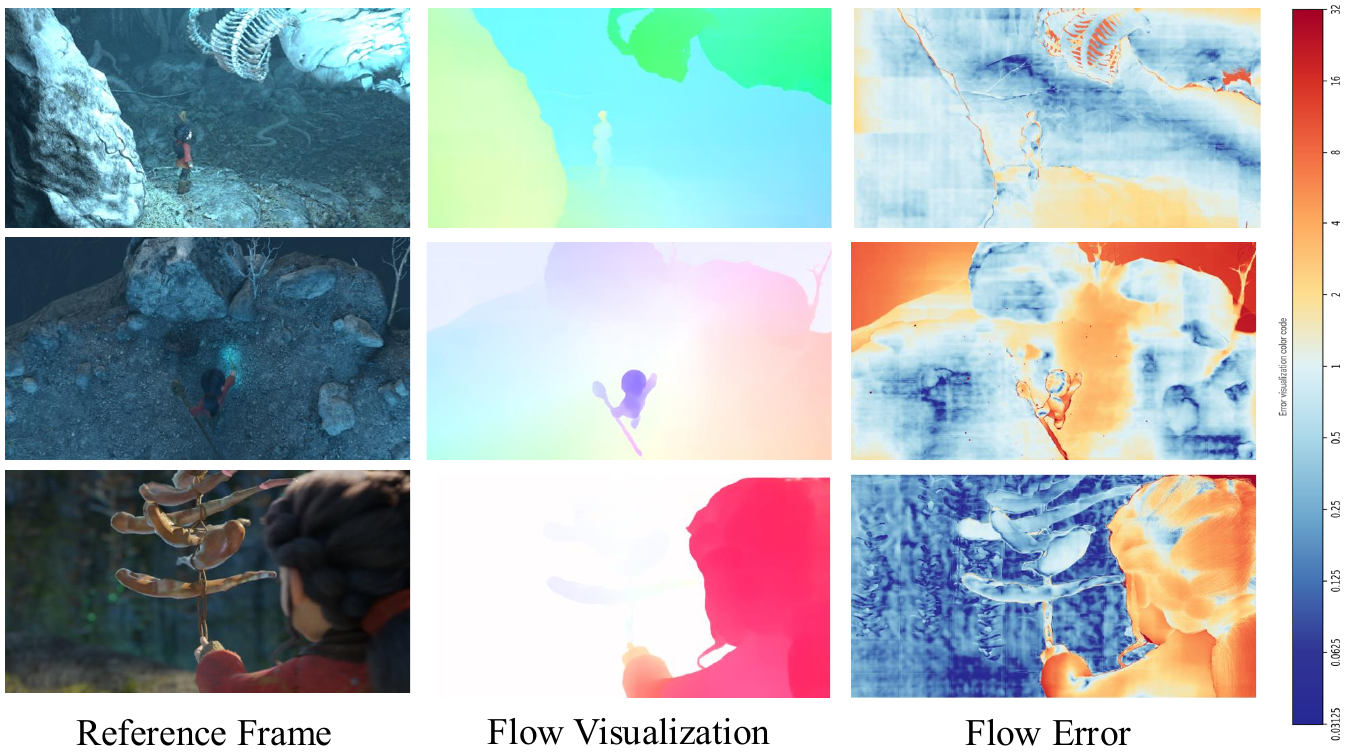}
\caption{\small \rev{Qualitative results on Spring Benchmark Dataset \cite{roco29Public2026}.}}\label{fig:qualitative_result}
\end{figure*}

\revlocation{challenge}
RoCo-Spring challenge benchmark provides high-resolution, high-detail dense optical flow evaluation~\citep{mehl2023spring,oei2026robustspring}. Our entry obtains $E_S=1.411$, $\Delta=3.328$, and $E_{R,\mathrm{proxy}}=4.739$, yielding RbS $1.1038$ after rounding, where $E_S$ denotes Spring EPE and $\Delta$ the mean clean-to-corrupted flow disagreement. \rev{ Figure \ref{fig:qualitative_result} show the qualitative result on Spring Benchmark dataset. 

The challenge-defined proxy is $E_{R,\mathrm{proxy}}=E_S+\Delta$, not a direct corrupted-input ground-truth error. Our balanced score is lower than seven of the twelve reference baselines in Table~\ref{tab:challenge}. Here, GMFlow results are quite interesting. GMFlow has lower clean EPE (0.945), disagreement (2.980), and proxy EPE (3.925). The clean-EPE gap is 0.466 pixels despite our additional 400k Spring fine-tuning updates. The GMFlow reference uses a Sintel checkpoint~\cite{oei2026robustspring}, whereas our entry is Spring-adapted and thus these are comparisons of configurations with different training histories.}

\rev{The breakdown in Appendix~\ref{app:breakdown} identifies different clean and corrupted-input limitations. Clean EPE is 7.193 for high-detail regions, 6.500 for unmatched pixels, and 5.887 for motions above 40 pixels. Disagreement is low for brightness (0.308) and saturation (0.415), but high for rain (13.034), spatter (10.097), snow (8.943), and frost (8.579). Thus, aggregate stability coexists with substantial sensitivity to particular corruptions. RobustSpring uses one specified severity per corruption~\cite{oei2026robustspring}. Appendix~\ref{app:breakdown} also discusses the public clean-input failure visualization and the distinction between EPE and the 1px error rate.}

\begin{table}[t]
\centering
\scriptsize
\setlength{\tabcolsep}{5pt}
\begin{tabular}{@{}lrrrr@{}}
\toprule
Method & RbS-Score & Spring EPE & Proxy EPE & $\Delta$\\
\midrule
SEA-RAFT \cite{wang2024sea} & 0.4584 & 0.363 & 3.323 & 2.960\\
MS-RAFT+ \cite{jahedi2023ms} & 0.6774 & 0.643 & 4.263 & 3.620\\
FlowFormer \cite{huang2022flowformer} & 0.7368 & 0.723 & 4.493 & 3.770\\
GMFlow \cite{xu2022gmflow} & 0.8015 & 0.945 & 3.925 & 2.980\\
GMA \cite{jiang2021learning} & 0.8704 & 0.914 & 4.944 & 4.030\\
\textbf{VSAI (RoCo-29, Method 1)} & \textbf{1.1038} & \textbf{1.411} & \textbf{4.739} & \textbf{3.328}\\
FlowNet2 \cite{ilg2017flownet} & 1.1914 & 1.040 & 8.050 & 7.010\\
RoCo-Spring Team baseline & 1.2375 & 1.493 & 5.853 & 4.360\\
RAFT \cite{teed2020raft} & 1.3336 & 1.476 & 7.116 & 5.640\\
M-FUSE (K) \cite{Mehl2023} & 1.7631 & 2.526 & 5.916 & 3.390\\
RAFT-3D (K) \cite{teed2021raft3d} & 1.9003 & 2.528 & 7.558 & 5.030\\
PWCNet \cite{sun2018pwc} & 1.9435 & 2.288 & 9.538 & 7.250\\
SPyNet \rev{\cite{spynet2017}} & 2.7976 & 4.162 & 8.452 & 4.290\\
\bottomrule
\end{tabular}
\caption{\small \textbf{RoCo-Spring optical flow result.} The reported columns compare the submitted entry with challenge reference methods. $\Delta=E_{R,\mathrm{proxy}}-E_S$ is shown for interpretation. All metrics are lower-is-better; EPE and disagreement are in pixels. Reference baselines are unranked challenge references, not competing team ranks.}
\label{tab:challenge}
\end{table}

\subsection{Ablations}

\formattedparagraph{\textit{(i)} Cross-Modal Fusion.}
\rev{Table~\ref{tab:fusion_ablation} reproduces the TTSFlow fusion ablation~\cite{Chanda_2026_CVPR}, without DPT or refinement.} Fusion reduces EPE from 3.52 to 3.02 on Things Clean, from 1.575 to 1.46 on Sintel Clean, and from 3.12 to 2.81 on Sintel Final, yielding a 10\% relative reduction on Final. The largest absolute improvement occurs for large displacements ($s_{40+}$) on Sintel Final, where EPE decreases from 19.37 to 16.99. Fusion improves all other reported motion bins except Sintel Clean $s_{10\text{--}40}$, where EPE increases slightly from 2.01 to 2.05. \rev{These historical results support the base fusion configuration under that protocol and are distinct from the DPT comparison below.}

\formattedparagraph{\textit{(ii)} Effects of DPT.}
Table~\ref{tab:dpt_ablation} compares the proposed pipeline with and without DPT after 360k training steps on Things, both with one refinement step. Adding DPT reduces EPE from 1.33 to 1.26 on Sintel Clean and from 2.73 to 2.37 on Sintel Final. Errors decrease across all reported motion bins, with the largest reduction in Sintel Final $s_{40+}$, from 16.69 to 14.05 EPE. With refinement count held fixed, these gains support the contribution of the DPT head to dense correspondence accuracy, particularly at large displacements. \textit{Additional experiments, ablations, and compute details are provided in the appendix}.

\begin{table}[h]
\scriptsize
\centering
\resizebox{\linewidth}{!}{
%\vspace{-0.3cm}
\begin{tabular}{l|c|cccc|cccc}
\hline
\multirow{2}{*}{Method}
& \multicolumn{1}{c|}{Things (val, clean)} 
& \multicolumn{4}{c|}{Sintel (train, clean)} 
& \multicolumn{4}{c}{Sintel (train, final)} \\
\cline{2-10}
& EPE
& EPE & s$_{0-10}$ & s$_{10-40}$ & s$_{40+}$ 
& EPE & s$_{0-10}$ & s$_{10-40}$ & s$_{40+}$ \\
\hline

(w/o) Fusion & 3.52
& 1.575 & 0.45 & 2.01 & 8.53
& 3.12 & 0.685 & 3.57 & 19.37 \\

(w/) Fusion & 3.02
& 1.46 & 0.39 & 2.05 & 7.74
& 2.81 & 0.64 & 3.36 & 16.99 \\

\hline
\end{tabular}
}
\caption{\small \rev{Fusion ablation reproduced from TTSFlow~\cite{Chanda_2026_CVPR}, Table 4, with 800k Things-stage updates following 200k Chairs updates, and no DPT or refinement. This is prior evidence for the base fusion design, not a new ablation of the submitted configuration.}}
\label{tab:fusion_ablation}
\end{table}

\begin{table*}[h]
\scriptsize
\centering
{
%\vspace{-0.3cm}
\begin{tabular}{l|cccc|cccc}
\hline
\multirow{2}{*}{Method}
& \multicolumn{4}{c|}{Sintel (train, clean)} 
& \multicolumn{4}{c}{Sintel (train, final)} \\
\cline{2-9}
& EPE & $s_{0-10}$ & $s_{10-40}$ & $s_{40+}$ 
& EPE & $s_{0-10}$ & $s_{10-40}$ & $s_{40+}$ \\
\hline

without DPT
& 1.33 & 0.40 & 1.61 & 7.26
& 2.73 & 0.62 & 3.16 & 16.69 \\

with DPT
& 1.26 & 0.34 & 1.52 & 7.25
& 2.37 & 0.55 & 2.92 & 14.05 \\

\hline
\end{tabular}
}
\caption{\small \rev{DPT comparison on the Sintel training split, Clean and Final passes, after 360k Things-stage updates following Chairs training. Both configurations use one refinement stage.}}
\label{tab:dpt_ablation}
\end{table*}

%% file: sec/5-conclusion.tex
\section{Conclusion, Limitations, and Broader Impact}
We presented a foundation-driven optical flow framework that reuses frozen visual and depth foundation representations for dense correspondence estimation. By combining semantic and geometric priors through lightweight fusion and global matching, our method estimates optical flow in a single forward pass without expensive multi-step test-time refinement, achieving strong cross-dataset generalization and competitive accuracy. The main limitation here is that foundation features are not optimized specifically for optical flow. Consequently, the model may provide inferior results under extreme motion, severe appearance changes, occlusions, or fine motion boundaries. Lightweight multiple refinement, adaptive correction, or multi-frame temporal reasoning could improve robustness while preserving the efficiency of the proposed design. Broadly, this work points toward reducing task-specific training and test-time scaling in dense visual perception. Such approaches may benefit robotics, video understanding, and 3D scene analysis, but optical flow failures in safety-critical settings still require careful validation and uncertainty-aware deployment.

%% file: sec/X-supp-w.tex
\supplementarytitlewithauthors{Foundation Model Driven Optical Flow\\---Appendix---}{}

\section{Public Spring and RobustSpring Breakdown}
\label{app:breakdown}
Tables~\ref{tab:springbreakdown} and~\ref{tab:corruptions} transcribe the EPE fields from the public RoCo-29 result\rev{~\cite{roco29Public2026}}. Clean Spring values are ground-truth errors. RobustSpring values are clean-to-corrupted prediction disagreements, $\Delta_c$, despite being displayed under an EPE heading on the server. The reported aggregate disagreement is 3.328. The challenge adds clean EPE 1.411 to obtain proxy EPE 4.739. The aggregate is retained as reported rather than recomputed by taking an unweighted mean of the corruption rows.

\begin{table}[H]
\caption{\textbf{Clean Spring EPE by evaluation subset.} Values are in pixels.}
\label{tab:springbreakdown}
\centering\small
\begin{tabular}{@{}lr@{\qquad\qquad}lr@{}}
\toprule
Subset & EPE & Subset & EPE\\
\midrule
Overall &1.411&Non-rigid&3.804\\
Low-detail&1.374&Not sky&1.406\\
High-detail&7.193&Sky&1.489\\
Matched&1.288&$s_{0-10}$&0.898\\
Unmatched&6.500&$s_{10-40}$&1.340\\
Rigid&1.094&$s_{40+}$&5.887\\
\bottomrule
\end{tabular}
\end{table}

\begin{table}[H]
\caption{\textbf{RobustSpring disagreement for all twenty corruptions.} $\Delta_c$ is in pixels. Lower values indicate less change in predicted flow. These are not corrupted-input ground-truth EPE values.}
\label{tab:corruptions}
\centering\small
\begin{tabular}{@{}lr@{\qquad\qquad}lr@{}}
\toprule
Corruption & $\Delta_c$ & Corruption & $\Delta_c$\\
\midrule
Brightness&0.308&JPEG compression&1.671\\
Contrast&3.213&Motion blur&1.538\\
Defocus blur&0.592&Pixelate&0.756\\
Elastic transform&1.189&Rain&13.034\\
Fog&1.382&Saturate&0.415\\
Frost&8.579&Shot noise&1.980\\
Gaussian blur&0.713&Snow&8.943\\
Gaussian noise&2.070&Spatter&10.097\\
Glass blur&0.604&Speckle noise&2.308\\
Impulse noise&2.882&Zoom blur&4.294\\
\bottomrule
\end{tabular}
\end{table}

\revlocation{robustnessdetail}
\begin{revision}
\paragraph{Corruption types and severity.}
Brightness (0.308 pixels) and saturation (0.415) produce relatively small changes in the estimated flow. Rain (13.034), spatter (10.097), snow (8.943), and frost (8.579) produce much larger disagreements. Blur sensitivity also varies: defocus gives 0.592 pixels, while zoom blur gives 4.294. These values identify settings in which the predictions are stable or unstable; they do not measure the corrupted predictions against ground truth. RobustSpring specifies one severity setting for each corruption~\cite{oei2026robustspring}. Table~\ref{tab:corruptions} reports those benchmark settings, not a severity sweep, so the results do not establish a trend with increasing severity.

\paragraph{Clean accuracy and a representative failure.}
On clean Spring, EPE rises from 1.411 overall to 7.193 in high-detail regions and 6.500 at unmatched pixels. The public result also reports a 34.890\% clean 1px error rate~\cite{roco29Public2026}. Mean EPE and this thresholded error describe different properties of the error distribution; neither alone establishes an advantage for large motion. The reported motion-stratified EPE is 0.898 for $s_{0-10}$, 1.340 for $s_{10-40}$, and 5.887 for $s_{40+}$.

A concrete clean-input example is the public RoCo-29 hero frame \texttt{hero\_00}, showing a running dog on textured ground, and its corresponding error visualization \texttt{flerrimg\_00}~\cite{roco29Public2026}. The error image contains warm-colored regions on the moving dog and across parts of the foreground, while parts of the upper background appear cooler. This example illustrates spatially nonuniform error on an uncorrupted pair; it is not a corruption-disagreement image. The displayed colorbar has no numerical tick labels, so no pixel-error magnitude is inferred from its colors. Corruption-specific spatial causes cannot be established from the aggregate values in Table~\ref{tab:corruptions} alone.
\end{revision}

\paragraph{Interpretation.}
The clean subset analysis separates the method's fine-detail and visibility limitations from its response to corrupted appearance. The corruption breakdown shows that favorable aggregate stability can coexist with pronounced sensitivity to particular disturbances. Because submitted-checkpoint evaluations with individual modules disabled are not reported here, this evidence is descriptive: it does not establish whether DINOv2 features, depth features, fusion, or the matching operator are responsible for any particular corruption response. Similarly, an unchanged but inaccurate prediction would have low disagreement; stability and clean error must therefore be read together.

\paragraph{Score calculation.}
Using the challenge's fixed constants and the displayed values gives
\begin{equation}
 \tfrac12\left(\frac{1.411}{0.9925}+\frac{1.411+3.328}{6.030}\right)
 =1.1037831\ldots\approx1.1038.
\end{equation}
\rev{The comparison counts the twelve listed reference baselines in Table~\ref{tab:challenge}. Our entry has a lower robustness-balanced score than seven of those references; this is neither a training-matched architectural comparison nor a claim about the final ranking of submitted teams.}

\section{Qualitative Comparisons}
\label{app:qualitative}
\begin{figure}[H]
\centering
\includegraphics[width=\linewidth]{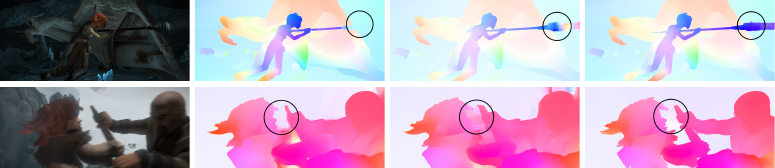}
\smallskip
\noindent\makebox[0.25\linewidth][c]{Input}\makebox[0.25\linewidth][c]{SEA-RAFT~\citep{wang2024sea}}\makebox[0.25\linewidth][c]{FlowSeek~\citep{Poggi_2025_ICCV}}\makebox[0.25\linewidth][c]{Ours}
\caption{\textbf{Sintel qualitative comparison.} Two examples from the Sintel benchmark~\citep{butler2012naturalistic}. The circles are annotations retained from the reported experiment. The visual comparisons illustrate local differences in predicted motion, without establishing a quantitative occlusion or boundary advantage.}
\label{fig:sintel}
\end{figure}

\begin{figure}[H]
\centering
\includegraphics[width=\linewidth]{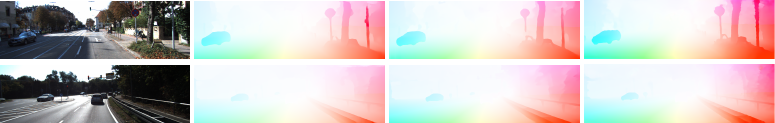}
\smallskip
\noindent\makebox[0.25\linewidth][c]{Input}\makebox[0.25\linewidth][c]{SEA-RAFT~\citep{wang2024sea}}\makebox[0.25\linewidth][c]{FlowSeek~\citep{Poggi_2025_ICCV}}\makebox[0.25\linewidth][c]{Ours}
\caption{\textbf{KITTI qualitative comparison.} Two examples from the KITTI training split~\citep{geiger2013vision}. These examples belong to the target-adapted evaluation and should not be interpreted as held-out test results.}
\label{fig:kitti}
\end{figure}

\revlocation{qualitative}
Figures~\ref{fig:sintel} and~\ref{fig:kitti} retain the qualitative comparisons accompanying the standard-benchmark experiments. Columns show the input frame and the predictions of SEA-RAFT, FlowSeek, and our method. Flow colors visualize estimated direction and magnitude; they are not error maps. These examples complement the quantitative tables but do not substitute for boundary-, visibility-, or corruption-specific evaluation.

%% file: sec/2-related-works.tex
\section{Related Work}\label{sec:relatedwork}
\revlocation{relatedwork}

%\vspace{-0.5cm}
\formattedparagraph{\textit{(i)} Learning-based Optical Flow.} 
Early works such as FlowNet \cite{dosovitskiy2015flownet} introduced supervised regression, while PWC-Net \cite{sun2018pwc} incorporated pyramidal cost volumes and warping to improve efficiency and accuracy. Subsequent approaches refined these ideas through multi-scale processing and architectural optimization. RAFT \cite{teed2020raft} established a dominant paradigm by formulating optical flow as iterative refinement over all-pairs correlations. Its recurrent update operator progressively corrects flow estimates, improving accuracy at the cost of increased inference time. SEA-RAFT \cite{wang2024sea} further optimized it, introducing probabilistic supervision and rigid-motion pretraining to improve efficiency and generalization, while \cite{kiefhaber2025removing} focused on speed and reduced memory needs. Despite these advances, RAFT and SEA-RAFT rely on multi-step test-time refinement to gain performance.

Transformer-based models such as GMFlow \cite{xu2022gmflow} and FlowFormer \cite{huang2022flowformer} shifted attention toward global reasoning. GMFlow presented optical flow as a global matching problem using attention and softmax normalization, yet requires supervised training and, in practice, benefits greatly from refinement. FlowFormer \cite{huang2022flowformer} extended this idea via cost-volume transformers to model long-range dependencies. \rev{Our pipeline inherits GMFlow's transformer, correlation-to-flow formulation, propagation, refinement, and convex upsampling. It replaces the CNN feature extraction path with fused foundation-model features while retaining supervised training of the downstream flow-estimation modules.}

\formattedparagraph{\textit{(ii)} Geometric Priors in Optical Flow.} Incorporating geometric cues into optical flow has long been an attractive direction \cite{sun2010secrets, wulff2015efficient}. Earlier methods relied on piecewise rigid motion assumptions \cite{vogel2013piecewise, yang2015dense}, whereas recent learning-based approaches inject depth priors directly into the network. Recently proposed FlowSeek \cite{Poggi_2025_ICCV} represents an interesting step in this direction. It integrates monocular depth foundation models \cite{yang2024depthv2} into a RAFT-style architecture and demonstrates improved cross-dataset generalization. \rev{FlowSeek uses depth features and depth-derived motion bases in a recurrent flow-estimation architecture. We instead fuse an intermediate Depth Anything V2 decoder feature with a separate DINOv2 feature stream before correlation-based matching. The foundation models remain fixed, while the downstream transformer and flow-estimation modules learn from optical-flow supervision. This distinction concerns where pretrained features enter the matching pipeline; it does not remove flow-specific learning.}

\formattedparagraph{\textit{(iii)} Vision Foundation Models for Dense Geometry.}
Vision foundation models have recently demonstrated remarkable transferability across tasks. Self-supervised models such as DINO \cite{caron2021emerging} and DINO-v2 \cite{oquab2023dinov2} learn dense, semantically meaningful representations that generalize across domains without explicit supervision. In parallel, depth foundation models such as Depth Anything \cite{yang2024depth} and its successors \cite{ yang2024depthv2} achieve strong cross-dataset generalization and produce boundary-aware geometric predictions. Dense prediction architectures such as the Dense Prediction Transformer (DPT) \cite{ranftl2021vision} further bridge the gap between transformer token representations and spatially structured outputs. DPT-style decoders reassemble intermediate transformer features into multi-scale dense maps, enabling tasks that require fine-grained spatial reasoning. \rev{Our visual DPT head decodes intermediate DINOv2 features before they are resampled to the global-matching and refinement grids. It is distinct from the pretrained DPT decoder inside Depth Anything V2. Figure~\ref{fig:overall_pipeline} and Appendix~\ref{app:featurepaths} specify both paths, including the exact depth-decoder feature used.}

\begin{revision}
\formattedparagraph{Direct predecessor and other pretrained architectures.}
TTSFlow~\cite{Chanda_2026_CVPR} is the direct predecessor of this submission: it already combines frozen DINOv2 and depth-foundation features, fusion, and global matching. Here we study DPT-based visual decoding, one higher-resolution refinement stage, and a Spring-adapted RoCo-Spring entry. The TTSFlow rows in Tables~\ref{tab:zeroshot_full} and~\ref{tab:sintel_trainedg}, and the fusion ablation in Table~\ref{tab:fusion_ablation}, reproduce the predecessor's reported results rather than new evaluations of the submitted configuration.

WAFT~\cite{wang2025waft} uses frozen pretrained input encoders, including Depth Anything V2 and DINOv3 variants, with high-resolution warping and recurrent updates instead of a cost volume. Thus, pretrained encoders are not unique to our method. Our configuration combines two distinct frozen feature streams and uses explicit global correlation followed by one local residual update. CroCo v2~\cite{weinzaepfel2023croco} pretrains a transformer through cross-view completion and adapts it to stereo matching and optical flow. Its cross-view pretraining differs from reusing the separately pretrained visual and monocular-depth models studied here. D4RT~\cite{zhang2026d4rt} uses a feed-forward video representation with spatiotemporal queries for dynamic 3D reconstruction and tracking. It provides a related example of unified geometric prediction, whereas our evaluated task is two-frame dense 2D optical flow. These are architectural comparisons; no training-matched accuracy comparison with WAFT, CroCo v2, or D4RT is reported here.
\end{revision}

\rev{\textbf{Baseline evaluation scope.} The FlowSeek KITTI row uses the mixed-dataset configuration described in the submitted experiment. It is not evidence from a training-matched comparison with our additional KITTI adaptation. The table therefore compares the reported configurations rather than isolating the effect of the architecture.}

%% file: sec/X-supp.tex
\section{Additional Ablation Studies}\label{app:ablations}
\revlocation{branchablation}
\formattedparagraph{\textit{(i)} Contribution of Individual Foundation Priors.}
We first analyze the relative contribution of the two foundation priors used in our framework 1) visual semantic representations from DINOv2 \cite{oquab2023dinov2} and geometric cues from the depth foundation model \cite{yang2024depthv2}. Table~\ref{tab:dino_depth_ablation} reports the results of selectively disabling individual components from a model trained on FlyingChairs and FlyingThings3D. 

Specifically, we evaluate three configurations. 
(i) \textbf{Depth-only}: the DINOv2 backbone and the cross-modal fusion module are removed, leaving only depth features for correspondence estimation. 
(ii) \textbf{DINO-only}: the depth encoder, depth projection module, and fusion module are removed, leaving only semantic features from DINOv2. 
(iii) \textbf{Depth + DINOv2}: the full representation combining semantic and geometric priors. 
\rev{All variants use the no-DPT, zero-refinement configuration. These component-removal results are distinct from independently training an otherwise matched single-branch model; they do not isolate the information intrinsically available in either encoder.}

\rev{The depth-only configuration gives 122.87/120.12 EPE on Sintel Clean/Final, compared with 36.15/36.42 for DINOv2-only and 1.46/2.81 for the joint configuration. The large changes show the sensitivity of this model configuration to branch removal. They do not establish that a depth representation lacks appearance information or identify which feature stream resolves a particular boundary or occlusion.}

\begin{figure*}[h]
\centering
\begin{subfigure}[h]{0.32\textwidth}
    \includegraphics[width=\linewidth]{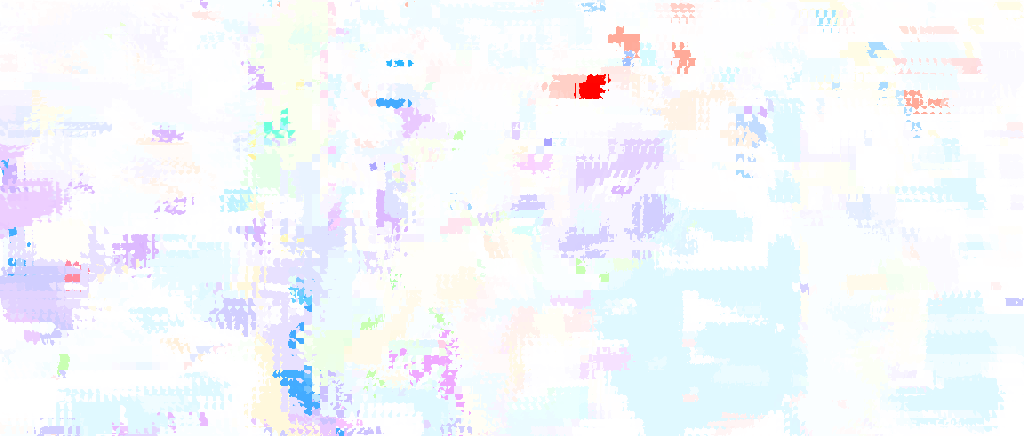}
    \centering{\small DINO}
\end{subfigure}
\begin{subfigure}[h]{0.32\textwidth}
    \includegraphics[width=\linewidth]{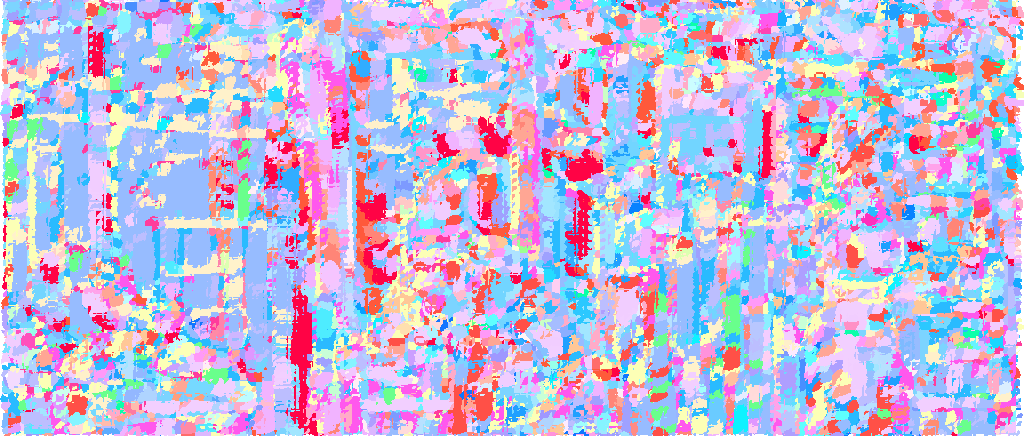}
    \centering{\small Depth}
\end{subfigure}
\begin{subfigure}[h]{0.32\textwidth}
    \includegraphics[width=\linewidth]{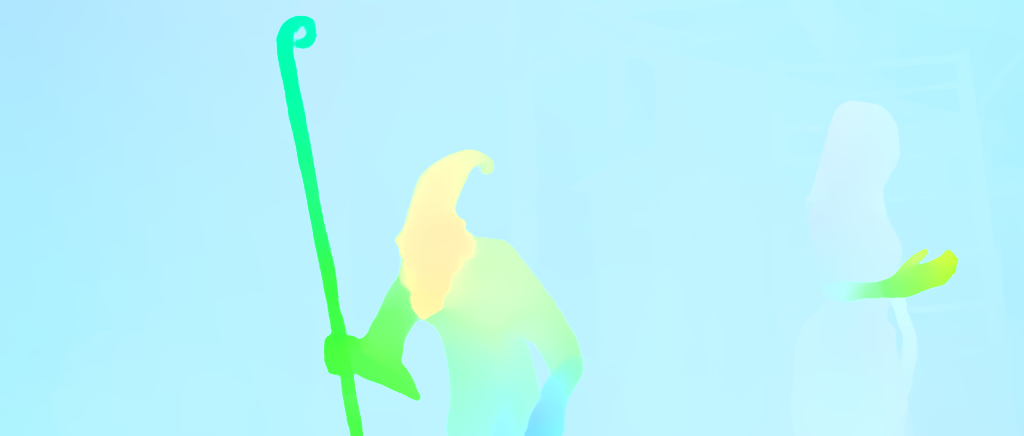}
    \centering{\small Depth + DINO}
\end{subfigure}
\caption{\rev{Predicted flow for the component-removal configurations in Table~\ref{tab:dino_depth_ablation}, without DPT or refinement. Colors encode flow, not ground-truth error.}}
\label{fig:feature_ablation}
\end{figure*}

\rev{The operational distinction matters when reading Table~\ref{tab:fusion_ablation}: its historical fusion comparison and the component removals in Table~\ref{tab:dino_depth_ablation} are different interventions. Their errors should not be treated as interchangeable measurements of a single ``DINOv2-only'' model.}

\begin{table*}[h]
\centering
\caption{\textbf{Ablation of DINOv2 and depth features.} \rev{Results on the Sintel training set for the reported component-removal configurations of a Chairs+Things model, without DPT or refinement. These are not a capacity-matched comparison of independently trained single-branch models.}}
\label{tab:dino_depth_ablation}
\resizebox{1.0\linewidth}{!}{
\begin{tabular}{|l|cccc|cccc|}
\hline
\multirow{2}{*}{Method} 
& \multicolumn{4}{c|}{Sintel (train, clean)} 
& \multicolumn{4}{c|}{Sintel (train, final)} \\
\cline{2-9}

& EPE & s$_{0-10}$ & s$_{10-40}$ & s$_{40+}$ 
& EPE & s$_{0-10}$ & s$_{10-40}$ & s$_{40+}$ \\
\hline

\textbf{Ours (Depth)}
& 122.87 & 103.32 & 152.01 & 197.83 
& 120.12 & 99.42 & 152.38 & 196.38 \\
\hline

\textbf{Ours (DINOv2)}
& 36.15 & 26.49 & 47.55 & 79.76 
& 36.42 & 25.33 & 49.20 & 87.06 \\
\hline

\textbf{Ours (Depth + DINOv2)}
& 1.46 & 0.39 & 2.05 & 7.74 
& 2.81 & 0.64 & 3.36 & 16.99 \\
\hline

\end{tabular}
}
\end{table*}

\smallskip
\formattedparagraph{\textit{(ii)} Effect of Depth Priors During Training.}
\revlocation{trainingablation}
\begin{revision}
Table~\ref{tab:depth_ablation_chairs} compares three configurations after 100k FlyingChairs training updates, without DPT or refinement. The joint configuration gives 0.87 EPE, compared with 1.77 without depth and 1.07 without DINOv2. The without-depth and joint values are also reported in TTSFlow~\cite{Chanda_2026_CVPR}; they are retained as prior evidence, rather than a new experiment on the submitted DPT/refinement model. Removing a feature branch also changes the fusion architecture, so the comparison does not separate feature content from downstream capacity. The evaluation split of these historical FlyingChairs ablations was not specified in the submitted record; the values are therefore not relabeled as validation-set results.
\end{revision}

\begin{table*}[h]
\centering
\caption{\rev{Three feature configurations after 100k FlyingChairs updates, without DPT or refinement. The 1.77 and 0.87 entries are retained from TTSFlow~\cite{Chanda_2026_CVPR}. The historical evaluation split is unspecified.}}
\label{tab:depth_ablation_chairs}
\begin{tabular}{|l|c|c|c|}
\hline
 & (w/o) Depth & (w/o) DINO & Depth+DINO  \\
\hline
\textbf{EPE} & 1.77 & 1.07 & 0.87  \\
\hline
\end{tabular}
\end{table*}

Additionally, we compare different backbone configurations for both DepthAnythingV2 and DINOv2 for the above model. \rev{Within each backbone comparison, the configurations are reported with the same FlyingChairs training settings. These comparisons change the pretrained backbone as well as its capacity.} For the DepthAnythingV2 variants, models are trained for 100k iterations, while the DINOv2 variants are trained for 50k iterations.

\begin{table}[!h]
\centering
\caption{\textbf{DepthAnythingV2 backbone comparison.} Results on the FlyingChairs dataset after training for 100k iterations.}
\label{tab:depthanything_backbone}
\begin{tabular}{|l|c|c|}
\hline
 & DepthAnythingV2 (Small) & DepthAnythingV2 (Base) \\
\hline
\textbf{EPE (FlyingChairs)} & 0.995 & 0.87 \\
\hline
\end{tabular}
\end{table}

\begin{table}[!h]
\centering
\caption{\textbf{DINOv2 backbone comparison.} Results on the FlyingChairs dataset after training for 50k iterations.}
\label{tab:dinov2_backbone}
\begin{tabular}{|l|c|c|}
\hline
 & DINOv2 (Small) & DINOv2 (Base) \\
\hline
\textbf{EPE (FlyingChairs)} & 1.091 & 1.056 \\
\hline
\end{tabular}
\end{table}

\section{Additional Experimental Results}

\subsection{Inference Time Comparison}
\revlocation{runtime}
We compare the inference speed of our method with several representative optical flow approaches. All measurements are conducted at the Sintel resolution using a single NVIDIA Quadro RTX 6000 GPU. Table~\ref{tab:inference_speed} reports the average inference time per image pair.
\rev{At 362 ms, our model is slower than every listed alternative, including GMFlow with one refinement (295 ms), RAFT with 32 updates (229 ms), and SEA-RAFT (S) with four updates (63 ms). One refinement stage describes our architecture; it does not imply lower inference latency.}

\begin{table}[H]
\centering
\caption{Inference time comparison of different optical flow methods.}
\label{tab:inference_speed}
\resizebox{1.0\linewidth}{!}{
\begin{tabular}{|l|c|c|c|c|c|c|}
\hline
 & RAFT & GMFlow & GMFlow & FlowSeek[T] & SEA-RAFT[S] & Ours \\
\hline
\textbf{\#Refine} & 32 & 0 & 1 & 4 & 4 & 1 \\
\hline
\textbf{Time (ms)} & 229 & 194 & 295 & 144 & 63 & 362 \\
\hline
\end{tabular}
}
\end{table}

\begin{revision}
The end-to-end timings do not isolate the costs of the two foundation-model branches, DPT decoding, fusion, matching, propagation, and upsampling. We therefore do not identify a dominant computational bottleneck from this table. The submitted timing record does not specify precision, timing batch size, warm-up, synchronization, or repetition count; the values are retained as reported rather than supplemented with an assumed protocol. Module-level profiling and a fully specified timing protocol would be needed for a reproducible efficiency attribution.
\end{revision}
% Original Table 13 (qualitative runtime adjectives, no measurements) removed;
% the measured timing table above is retained unchanged.

\subsection{\rev{KITTI Training-Split DPT Ablation}}
\revlocation{kittiappendix}

\begin{revision}
Table~\ref{tab:kitti_full_train} reports the KITTI training-split comparison after 140k KITTI fine-tuning updates, with one refinement stage. The DPT configuration obtains 1.29 EPE and 4.29\% F1-all, compared with 1.337 and 4.74\% without DPT. These are target-adapted training-split results. The 140k comparison is reported separately from the 200k KITTI setting in Table~\ref{tab:sintel_trainedg}, whose DPT result is 1.248 EPE and 4.01\% F1-all.
\end{revision}

\begin{table}[h]
\centering
\setlength{\tabcolsep}{4pt}
\caption{\textbf{Performance on the KITTI training set} \rev{after 140k KITTI fine-tuning updates, following Chairs, Things, and mixed TSKH training. Both configurations use one refinement stage; this is separate from the 200k setting in Table~\ref{tab:sintel_trainedg}.}}
\label{tab:kitti_full_train}
\begin{tabular}{|l|c|c|c|}
\hline
Method & \#refine & KITTI EPE & KITTI F1-all \\
\hline

\textbf{Ours (without DPT)} 
& 1
& 1.337 & 4.74 \\
\hline

\textbf{Ours (with DPT)} 
& 1
& \textbf{1.29} & \textbf{4.29} \\
\hline

\end{tabular}
\end{table}

\subsection{Extended Training On Chairs+Things}

\revlocation{extendedtraining}
\begin{revision}
Table~\ref{tab:zeroshot_full-1} compares the reported 360k and 560k Things-stage checkpoints after Chairs training. Longer training reduces Things validation EPE from 3.28 to 2.95 and Sintel training-split Clean EPE from 1.26 to 1.16. Sintel Final EPE increases from 2.37 to 2.42. The change therefore improves two reported evaluations while slightly worsening the third; it does not support a uniform generalization gain.
\end{revision}

\begin{table}[H]
\centering
\caption{\small \rev{Generalization after Chairs training and the indicated Things-stage training duration. The 560k checkpoint improves Things and Sintel Clean EPE relative to the 360k checkpoint, while Sintel Final EPE increases from 2.37 to 2.42.}}
%\vspace{\baselineskip}
\label{tab:zeroshot_full-1}
\resizebox{1.0\linewidth}{!}{
\begin{tabular}{|l|c|c|cccc|cccc|}
\hline
\multirow{2}{*}{Method} & \multirow{2}{*}{\#refine} 
& \multicolumn{1}{c|}{Things (val)} 
& \multicolumn{4}{c|}{Sintel (train, clean)} 
& \multicolumn{4}{c|}{Sintel (train, final)} \\
\cline{3-11}
& & EPE
& EPE & s$_{0-10}$ & s$_{10-40}$ & s$_{40+}$ 
& EPE & s$_{0-10}$ & s$_{10-40}$ & s$_{40+}$ \\
\hline

RAFT \cite{teed2020raft} & 32 & 4.25 
& 1.43 & 0.33 & 1.54 & 9.04 
& 2.71 & 0.51 & 2.98 & 17.62 \\
\hline

GMFlow \cite{xu2022gmflow} & 0 & 3.48 
& 1.50 & 0.46 & 1.77 & 8.26 
& 2.96 & 0.72 & 3.45 & 17.70 \\
\hline

GMFlow \cite{xu2022gmflow} & 1 & 2.80 
& 1.08 & 0.30 & 1.25 & 6.26 
& 2.48 & 0.51 & 2.81 & 15.76 \\
\hline

SEA-RAFT (S) \cite{wang2024sea} & 4 & - 
& 1.27 & - & - & - 
& 4.32 & - & - & - \\
\hline

FlowSeek (T) \cite{Poggi_2025_ICCV} & 4 & 3.94
& 1.12 & 0.23 & 1.27 & 7.12 
& 2.52 & 0.45 & 2.78 & 16.60 \\
\hline

\textbf{Ours (360k) } & \textbf{1}
& \textbf{3.28} 
& \textbf{1.26} & 0.34 & 1.52 & 7.25 
& \textbf{2.37} & 0.55 & 2.92 & \textbf{14.05} \\
\hline

\textbf{Ours (560k)} & \textbf{1}
& \textbf{2.95} 
& \textbf{1.16} & 0.31 & 1.30 & 6.87 
& \textbf{2.42} & 0.52 & 2.76 & \textbf{15.09} \\
\hline

\end{tabular}
}
\end{table}

% REV O21: replace unsupported mechanistic analysis, original PDF lines 556--593.
\section{\rev{Scope of the Representation Evidence}}\label{app:evidencescope}
\revlocation{evidencescope}
\begin{revision}
The experiments answer different questions. Table~\ref{tab:dpt_ablation} compares DPT configurations with refinement count held at one. Table~\ref{tab:fusion_ablation} reproduces prior fusion evidence for the no-DPT, zero-refinement TTSFlow configuration. Table~\ref{tab:dino_depth_ablation} reports sensitivity to removing components, whereas Table~\ref{tab:depth_ablation_chairs} concerns configurations trained with different feature inputs. These results support the evaluated joint configurations, but do not establish a particular semantic-versus-geometric division of labor or the causal effect of freezing. No frozen-versus-fine-tuned comparison or module-specific corruption ablation is reported. The complete system includes a trained transformer after feature fusion, and its measured latency is reported independently of the number of refinement stages.
\end{revision}

%% file: sec/Y-implementation.tex
% Kumar: Added after the reviewers comments
\clearpage
\section{\rev{Implementation Details and Source-Derived Inventory}}\label{app:implementation}
\revlocation{implementation}
\begin{revision}
This appendix documents the public FFlow implementation. The architecture and parameter counts below are derived from the released source. A source establishes its forward computation and configuration values; it is not a record of which code version, initialization, or evaluation command produced an earlier checkpoint. We distinguish this evidence from the experimental settings reported in the submitted manuscript wherever they differ.
\end{revision}

\subsection{\rev{Exact Feature Paths}}\label{app:featurepaths}
\revlocation{featurepaths}
\begin{revision}
\paragraph{Visual branch.}
The RGB inputs are normalized and bilinearly resized to $448\times448$ for DINOv2-S. With patch size 14, each of blocks 3, 6, 9, and 12 supplies $32\times32$ patch tokens with 384 channels, excluding the class token. The visual DPT head applies $1\times1$ projections to 48, 96, 192, and 384 channels. Its reassembly operations have spatial factors $8,4,2,1$, yielding grids of $256^2,128^2,64^2,32^2$, respectively. Each map is projected to 128 channels by a $3\times3$ convolution, then processed by the DPT fusion blocks. The finest output, $256\times256\times128$, is bilinearly resized independently to $H/8\times W/8$ and $H/4\times W/4$. In code these operations are in \texttt{DinoV2SFeature} and the visual \texttt{DPTHead}; the $1/4$ feature is not an interpolation of raw native $1/8$ DINO tokens.

\paragraph{Depth branch.}
Depth Anything V2-B uses its own DINOv2-B encoder and pretrained depth DPT decoder. The input is resized to $518\times518$, giving a $37\times37$ patch grid with 768-dimensional encoder features. The feature used for flow is \texttt{path\_1}, the output of \texttt{depth\_head.scratch.refinenet1}, before \texttt{output\_conv1}. In the released implementation its native size is $296\times296\times128$: the final fusion block upsamples the $148\times148$ level by two. This is distinct from the scalar depth output, which is computed but not consumed by the flow estimator. The selected feature is resized to $H\times W$ and passed through three $3\times3$, stride-2 convolutions, with channel widths $128\to192\to256\to128$. The second and third outputs supply the $1/4$ and $1/8$ grids. A separate $1\times1$ projection reduces the $1/4$ output from 256 to 128 channels. Image/feature resizing in these paths uses bilinear interpolation with \texttt{align\_corners=False}.

\paragraph{Fusion and learned matching.}
Each scale concatenates two 128-channel features. Fusion consists of a bias-free $3\times3$ convolution at 256 channels, instance normalization and a leaky ReLU, followed by two residual blocks, each containing two biased $3\times3$ convolutions, instance normalization, and ReLU activations. A biased $1\times1$ projection gives 128 output channels. The fusion and output projection share weights across frames and scales. The six-block matching transformer uses single-head self- and cross-attention, width 128, and feed-forward expansion factor four. The cross-attention feed-forward network consumes the concatenated source and message, so its widths are $256\to1024\to128$. The spatial window partitions are $2\times2$ at the coarse scale and $8\times8$ at the fine scale, with alternating shifted blocks. Positional encoding is added before the transformer.

For global propagation, the released code uses $Q=\psi_q(F_1)$ and $K=\psi_k(Q)$, including the learned projections in the attention affinity. Fine-scale propagation uses independently projected queries and keys from the first-frame feature, within a $3\times3$ neighborhood. The convex-upsampling mask predictor takes the 128-channel first-frame feature concatenated with two flow channels and applies convolutions $130\to256$ (kernel 3) and $256\to144$ (kernel 1). For each of the $4\times4$ subpixels, a softmax over nine neighboring flow vectors supplies the convex weights. Flow-vector values are multiplied by four before aggregation.
\end{revision}

\subsection{\rev{Frozen Parameters, Gradient Paths, and Counts}}\label{app:parameters}
\revlocation{parameters}
\begin{revision}
Both pretrained foundation models are frozen. The public constructor loads DINOv2-S from the Hugging Face model and Depth Anything V2-B from the supplied depth checkpoint. The downstream transformer, depth projections, fusion, propagation projections, and upsampling predictor participate in flow-loss backpropagation. The matching transformer, propagation, and convex upsampling follow GMFlow; the depth projections and feature fusion follow the TTSFlow design. Architectural provenance does not, by itself, imply initialization from a pretrained GMFlow optical-flow checkpoint. The released constructor does not load such a checkpoint; subsequent \texttt{--resume} loads the specified FFlow state.

\paragraph{Visual-DPT gradient status.}
The public \nolinkurl{FFlow.forward} wraps the entire \nolinkurl{dino_extract_feature} call in \nolinkurl{torch.no_grad()}. That call includes the added visual DPT head as well as the frozen DINOv2 encoder. Thus, although the visual DPT parameters have \texttt{requires\_grad=True}, the released forward path does not produce flow-loss gradients for them. Their initial values come from the layer constructors unless overwritten by a resumed FFlow state; no separate pretrained visual-DPT checkpoint is loaded there. This distinction must be resolved against the training snapshot before attributing the reported DPT comparison to a flow-trained visual decoder. Merely counting parameters with \texttt{requires\_grad=True} would overstate the gradient-connected portion of this release.
\end{revision}

\begin{table}[H]
\centering\small
\begin{revision}
\caption{\rev{Source-derived parameter inventory of the released two-scale configuration. Counts include biases and embedding/normalization parameters; shared modules are counted once. ``No flow gradient'' describes the released computation, not a reconstruction of historical training. No experimental checkpoint was loaded to obtain these counts.}}
\label{tab:parameter_inventory}
\begin{tabular}{@{}lrl@{}}
\toprule
Module & Parameters & Status in released flow path\\
\midrule
DINOv2-S encoder &22,056,576&Frozen\\
Added visual DPT head &4,124,576&No flow gradient; flag true\\
Depth Anything V2-B encoder &86,580,480&Frozen\\
Depth Anything V2 DPT decoder &10,890,305&Frozen\\
Depth downsampling &959,040&Gradient-connected\\
Quarter-scale depth projection &32,896&Gradient-connected\\
Fusion residual network &2,950,144&Gradient-connected\\
Shared fusion output projection &32,896&Gradient-connected\\
Matching transformer &3,150,336&Gradient-connected\\
Flow propagation projections &33,024&Gradient-connected\\
Convex-upsampling predictor &336,784&Gradient-connected\\
\midrule
Total registered &131,147,057&\\
Frozen by parameter flags &119,527,361&\\
Flagged \texttt{requires\_grad=True} &11,619,696&Includes visual DPT\\
Flow-gradient-connected subset &7,495,120&Excludes visual DPT\\
\bottomrule
\end{tabular}
\end{revision}
\end{table}

\begin{revision}
The DINOv2-S inventory uses the published model configuration~\cite{dinov2SmallConfig}, including its stored $37\times37$ positional grid, even though the visual branch is evaluated on a $32\times32$ patch grid after input resizing. The depth decoder count includes 92,289 parameters in the scalar-depth output head, whose prediction is unused for flow. The counts are therefore registered model parameters, not a measure of inference operations, memory use, or latency. The accompanying source-shape audit records the arithmetic used for this inventory.
\end{revision}

\subsection{\rev{Training Stages and Cropping}}\label{app:training}
\revlocation{trainingrecord}
\begin{revision}
Table~\ref{tab:training_release} combines the stage durations reported in the manuscript with the settings specified by the released \texttt{run\_all.py}. Batch sizes in that launcher are global; the distributed path divides the value across GPUs. The manuscript reports training on two NVIDIA RTX 6000 GPUs. Saved run arguments are needed to establish that a particular submitted checkpoint used every launcher setting.
\end{revision}

\begin{table}[H]
\centering\small
\begin{revision}
\caption{\rev{Reported stage durations and released launcher settings. Durations denote updates within each stage, rather than cumulative updates over all stages. The launcher uses step limits one larger than the displayed round values and names checkpoints at the round values. The KITTI crop discrepancy is recorded explicitly below.}}
\label{tab:training_release}
\setlength{\tabcolsep}{5pt}
\begin{tabular}{@{}lrrll@{}}
\toprule
Stage & Updates & Global batch & Crop $(H\times W)$ & Peak learning rate\\
\midrule
FlyingChairs &200k&8&$384\times512$&$4\times10^{-4}$\\
FlyingThings3D &360k&8&$384\times768$&$2\times10^{-4}$\\
TSKH mixture &320k&4&$320\times896$&$2\times10^{-4}$\\
KITTI &200k&4&$320\times1152$ (code)&$2\times10^{-4}$\\
Spring &400k&4&$544\times960$&$1\times10^{-4}$\\
\bottomrule
\end{tabular}
\end{revision}
\end{table}

\begin{revision}
\paragraph{Recorded discrepancy and initialization chain.}
The submitted manuscript specifies a KITTI crop of $320\times1192$, whereas the released launcher specifies $320\times1152$. Both values are documented rather than silently replacing the experimental record with the launcher. The crop used for the reported checkpoint requires confirmation from saved run arguments. The launcher initializes Things from the 200k Chairs checkpoint, TSKH from the 360k Things checkpoint, KITTI from the 320k TSKH checkpoint, and Spring from the 200k KITTI checkpoint. Named stage checkpoints contain model weights, while \texttt{checkpoint\_latest.pth} additionally contains optimizer, epoch, and step state. The latter can resume an interrupted stage; it is not equivalent to the model-only initialization used in the listed stage chain.

\paragraph{Optimizer and sampling.}
The released defaults use AdamW with weight decay $10^{-4}$, gradient-norm clipping at 1, and seed 326. The OneCycle learning-rate scheduler uses the stage's peak rate, a 5\% increasing portion, cosine decay, and no momentum cycling. The Things stage combines its Clean and Final renders. The TSKH dataset concatenates $100\times$ Sintel Clean, $100\times$ Sintel Final, $200\times$ KITTI, $5\times$ HD1K, and one copy of Things Clean. These are dataset-repeat factors, not per-step batch quotas.

\paragraph{Paired random crops.}
The training augmentors use random paired crops, not center crops. The same sampled crop is applied to the two images, the flow, and applicable validity information. Spatial resizing also rescales horizontal and vertical flow components. Dense-data crops sample a valid origin after spatial augmentation; the sparse KITTI augmentor samples with vertical/horizontal margins of 20/50 pixels and clips to valid origins, so its distribution is not the same uniform crop distribution. The Spring augmentor has its own padding, resizing, and random-crop procedure. In the released dataset definitions, the log$_2$ resize ranges are $[-0.1,1.0]$ for Chairs, $[-0.4,0.8]$ for Things, $[-0.2,0.6]$ for TSKH's Things/Sintel components, $[-0.3,0.5]$ for its KITTI component, $[-0.5,0.2]$ for HD1K, $[-0.2,0.4]$ for KITTI-only adaptation, and $[-1.0,-0.8]$ for Spring. The augmentors also enforce crop-size feasibility; these ranges alone do not specify the realized scale distribution. KITTI-only adaptation disables flips.

\paragraph{Evaluation record.}
The main tables distinguish cross-dataset evaluation from adaptation to datasets subsequently evaluated on their training splits. Predictions, not model weights, are submitted to the challenge server. The public entry identifies the reported benchmark outputs, but the supplied release does not by itself establish the challenge checkpoint hash, inference resizing/padding command, or final unpadding procedure used for that entry. We do not infer those settings from training crop sizes or from the feature-branch input sizes.
\end{revision}

\subsection{\rev{Supervised Predictions and Loss Normalization}}\label{app:supervision}
\revlocation{supervisiondetail}
\begin{revision}
In training mode, the released two-scale forward path appends four full-resolution predictions in this order:
\begin{center}
\begin{tabular}{@{}cllc@{}}
\toprule
$k$ & Native prediction & Full-resolution conversion & Weight\\
\midrule
1 & $1/8$ global matching & Bilinear $\times8$ &0.729\\
2 & $1/8$ global propagation & Bilinear $\times8$ &0.81\\
3 & $1/4$ residual matching & Bilinear $\times4$ &0.9\\
4 & $1/4$ local propagation & Learned convex $\times4$ &1\\
\bottomrule
\end{tabular}
\end{center}
Each conversion multiplies displacement components by the corresponding spatial factor. The bilinear flow upsampling uses \texttt{align\_corners=True}; this differs from the image/feature-resizing setting described above. Only the final prediction is returned as the output flow at inference.

For batch size $B$ and model-input size $H\times W$, define
$m_{bxy}=\mathbf{1}\{\mathrm{valid}_{bxy}\geq0.5\}
\mathbf{1}\{\|\mathbf{V}_{gt,bxy}\|_2<400\}$.
The implemented loss is
\begin{equation}
\mathcal{L}_{\mathrm{flow}}=
\sum_{k=1}^{4}\gamma^{4-k}\frac{1}{2BHW}
\sum_{b=1}^{B}\sum_{x=1}^{W}\sum_{y=1}^{H}
m_{bxy}\left\|\mathbf{V}^{(k)}_{bxy}-\mathbf{V}_{gt,bxy}\right\|_1,
\qquad\gamma=0.9.
\label{eq:loss_exact}
\end{equation}
The denominator includes all spatial positions and both flow components; it is not the number of valid pixels. Multiple supervised predictions arise from matching and propagation at two resolutions, even though there is only one higher-resolution refinement stage.
\end{revision}